\documentclass[10pt,twocolumn,letterpaper]{article}

\usepackage{cvpr}              

\makeatletter
\@namedef{ver@everyshi.sty}{}
\makeatother
\usepackage{tikz}

\usepackage{graphicx}
\usepackage{amsmath}
\usepackage{amssymb}
\usepackage{booktabs}

\usepackage{multirow}
\usepackage{makecell}

\usepackage{algorithm}
\usepackage{algpseudocode}

\usepackage{pgfplots}
\usepackage{pgfplotstable}
\renewcommand\thefootnote{}
\usepackage{enumitem}
\newenvironment{tight_itemize}{
\begin{itemize}[leftmargin=20pt]
  \setlength{\topsep}{0pt}
  \setlength{\itemsep}{0pt}
  \setlength{\parskip}{0pt}
  \setlength{\parsep}{0pt}
}{\end{itemize}}

\definecolor{color1}{rgb}{0.98, 0.81, 0.69}
\definecolor{color2}{rgb}{0.55, 0.71, 0.0}
\definecolor{color3}{rgb}{1.0, 0.6, 0.4}
\definecolor{color4}{rgb}{0.29, 0.59, 0.82}

\pgfplotscreateplotcyclelist{mycolor}{
{fill=color1!80, draw=color1},
{fill=color2!80, draw=color1},
{fill=color3!80, draw=color1},
{fill=color4!80, draw=color1},
}

\pgfplotscreateplotcyclelist{mycolor1}{
{fill=color1!80, draw=color2},
{fill=color2!80, draw=color2},
{fill=color3!80, draw=color2},
{fill=color4!80, draw=color2},
}

\pgfplotsset{
    discard if/.style 2 args={
        x filter/.code={
            \ifdim\thisrow{#1} pt=#2pt
                
            \fi
        }
    },
    discard if not/.style 2 args={
        x filter/.code={
            \ifdim\thisrow{#1} pt=#2pt
            \else
                
            \fi
        }
    }
}

\usepackage[pagebackref,breaklinks,colorlinks]{hyperref}

\usepackage[capitalize]{cleveref}
\crefname{section}{Sec.}{Secs.}
\Crefname{section}{Section}{Sections}
\Crefname{table}{Table}{Tables}
\crefname{table}{Tab.}{Tabs.}

\def\cvprPaperID{5782} 
\def\confName{CVPR}
\def\confYear{2022}

\usepackage[accsupp]{axessibility}  

\begin{document}

\title{Decoupled Multi-task Learning with Cyclical Self-Regulation for Face Parsing}

\author{Qingping Zheng$^1$, Jiankang Deng$^2$, Zheng Zhu$^3$, Ying Li$^1$, Stefanos Zafeiriou$^{2,4}$\\
\small $^1$Northwestern Polytechnical University, $^2$ Huawei, $^3$Tsinghua University, $^4$Imperial College London\\
{\tt\small zhengqingping2018@mail.nwpu.edu.cn, jiankangdeng@gmail.com, zhengzhu@ieee.org} \\
{\tt\small lybyp@nwpu.edu.cn, s.zafeiriou@imperial.ac.uk}
}
\maketitle

\begin{abstract}

This paper probes intrinsic factors behind typical failure cases (\eg spatial inconsistency and boundary confusion) produced by the existing state-of-the-art method in face parsing. To tackle these problems, we propose a novel Decoupled Multi-task Learning with Cyclical Self-Regulation (DML-CSR) for face parsing. Specifically, DML-CSR designs a multi-task model which comprises face parsing, binary edge, and category edge detection. 
These tasks only share low-level encoder weights without high-level interactions between each other, enabling to decouple auxiliary modules from the whole network at the inference stage.
To address spatial inconsistency, we develop a dynamic dual graph convolutional network to capture global contextual information without using any extra pooling operation.
To handle boundary confusion in both single and multiple face scenarios, we exploit binary and category edge detection to jointly obtain generic geometric structure and fine-grained semantic clues of human faces.
Besides, to prevent noisy labels from degrading model generalization during training, cyclical self-regulation is proposed to self-ensemble several model instances to get a new model and the resulting model then is used to self-distill subsequent models, through alternating iterations.
\footnote{This work is done when Qingping Zheng is an intern at Huawei.} 
\setcounter{footnote}{0}
\renewcommand\thefootnote{\arabic{footnote}}
Experiments show that our method achieves the new state-of-the-art performance on the Helen, CelebAMask-HQ, and Lapa datasets. 
The source code is available at \url{https://github.com/deepinsight/insightface/tree/master/parsing/dml_csr}.
\end{abstract}

\section{Introduction}

Face parsing, as a fine-grained semantic segmentation task, intends to assign a pixel-wise label for each facial component, \eg, eyes, nose, and mouth. The detailed analysis of semantic facial parts is essential in many high-level applications, such as face swapping \cite{nirkin2019fsgan}, face editing \cite{CelebAMask-HQ}, and facial makeup \cite{ou2016beauty}. Benefit from the learning capacity of deep Convolutional Neural Networks (CNNs) and the labor effort put in pixel-level annotations \cite{helen, lapa, CelebAMask-HQ}, methods based on Fully Convolutional Networks (FCNs) \cite{ehanet, multi_objective, icnn, cnn_cascade, lius, Zhou2017FacePV, guo2018, te2020edge, lin2021roi} have achieved a promising performance on the fully supervised face parsing. Nevertheless, the local characteristic of the convolutional kernel prevents FCNs from capturing global contextual information \cite{Merget_2018_CVPR}, which is crucial for semantically parsing facial components in an image.

\begin{figure}
\captionsetup[subfigure]{labelformat=empty}
\centering
\subfloat[Image]{
    \begin{minipage}[t]{0.185\linewidth}
    \includegraphics[width=1.6cm, height=1.6cm]{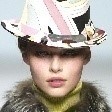}
    \\ \\[-2.7ex]
    \includegraphics[width=1.6cm, height=1.6cm]{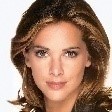}
    \\ \\[-2.7ex]
    \subfloat[Image]{\includegraphics[width=1.6cm, height=1.6cm]{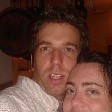}}
    \\ \\
    \includegraphics[width=1.6cm, height=1.6cm]{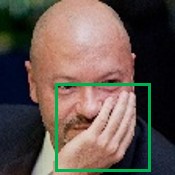}
    \end{minipage}
}
\subfloat[GT]{
    \begin{minipage}[t]{0.185\linewidth}
    \includegraphics[width=1.6cm, height=1.6cm]{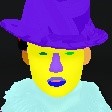}
    \\ \\[-2.7ex]
    \includegraphics[width=1.6cm, height=1.6cm]{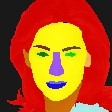}
    \\ \\[-2.7ex]
    \subfloat[GT]{\includegraphics[width=1.6cm, height=1.6cm]{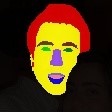}}
    \\ \\
    \includegraphics[width=1.6cm, height=1.6cm]{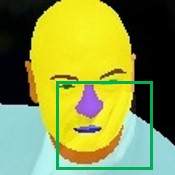}
    \end{minipage}
}
\subfloat[Image]{
    \begin{minipage}[t]{0.185\linewidth}
    \includegraphics[width=1.6cm, height=1.6cm]{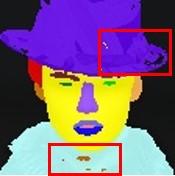}
    \\ \\[-2.7ex]
    \includegraphics[width=1.6cm, height=1.6cm]{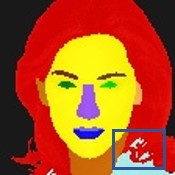}
    \\ \\[-2.7ex]
    \subfloat[EAGRNet]{\includegraphics[width=1.6cm, height=1.6cm]{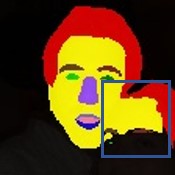}}
    \\ \\
    \includegraphics[width=1.6cm, height=1.6cm]{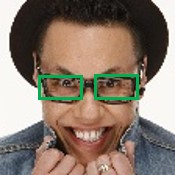}
    \end{minipage}
}
\subfloat[GT]{
    \begin{minipage}[t]{0.185\linewidth}
    \includegraphics[width=1.6cm, height=1.6cm]{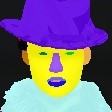}
    \\ \\[-2.7ex]
    \includegraphics[width=1.6cm, height=1.6cm]{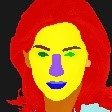}
    \\ \\[-2.7ex]
    \subfloat[Ours]{\includegraphics[width=1.6cm, height=1.6cm]{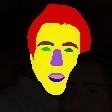}}
    \\ \\
    \includegraphics[width=1.6cm, height=1.6cm]{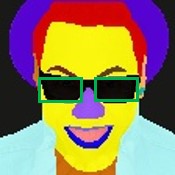}
    \end{minipage}
}
\vspace{-2.5mm}
\caption{The first three rows show typical failure cases of spatial inconsistency and boundary confusion when applying EARGNet \cite{te2020edge} to face parsing. The last row displays noisy labels on the training datasets.}
\vspace{-6mm}
\label{fig:failure_case_of_fcn}
\end{figure}

To address this issue, most of the region-based face parsing methods \cite{multi_objective, cnn_cascade, Zhou2017FacePV} integrate CNN features into variant CRFs to learn global information. However, these methods do not consider the correlation among various objects. To this end, Te \etal \cite{te2020edge} proposes the EAGRNet method to model a region-level graph representation over a face image by propagating information across all vertices on the graph. Even though EAGRNet enables reasoning over non-local regions to get global dependencies between distant facial components and achieves state-of-the-art performance, it still faces the problems of spatial inconsistency and boundary confusion. 
In EAGRNet, PSP module \cite{psp} adopts an average pooling layer \cite{Long_2015_CVPR} to capture the global context prior, leading to an inconsistent spatial topology. 
Moreover, EAGRNet integrates additional clues of binary edges into context embedding to improve the parsing results. However, it is hard for EAGRNet to handle boundaries between highly irregular facial parts (\eg hair and cloth in \figureautorefname{ \ref{fig:failure_case_of_fcn}}) and distinguish clear boundaries between different face instances in the crowded scenarios (multi-faces in \figureautorefname{ \ref{fig:failure_case_of_fcn}}). 

Besides, learning a reliable model for face parsing requires accurate pixel-level annotations. Nonetheless, there inevitably exist careless manual labeling errors on the training dataset as shown in the last row of \figureautorefname{ \ref{fig:failure_case_of_fcn}}. 
Te \etal \cite{te2020edge} employ the traditional fully supervised learning scheme to train EAGRNet, failing to locate label noise because all pixels in the ground truth are processed equally. Notably, overlooking such incomplete annotations restricts the model generalization and prevents the performance from increasing to a higher level.

In this paper, we propose an end-to-end face parsing method, which is based on Decoupled Multi-task Learning with Cyclical Self-Regulation (DML-CSR). Specifically, given an input of facial image, the ResNet-101 \cite{resnet101} pre-trained on ImageNet is taken as the backbone to extract features from different levels. Afterwards, our multi-task model consists of three tasks, namely face parsing, binary edge detection, and category edge detection. These tasks share low-level weights from the backbone but do not have high-level interactions. Therefore, our multi-task learning approach can detach additional edge detection tasks from face parsing at the inference stage. To tackle spatial inconsistency raised by the pooling operation, we develop a Dynamic Dual Graph Convolutional Network (DDGCN) in the face parsing branch to capture long-range contextual information. The proposed DDGCN contains no extra pooling operation and it can dynamically fuse the global context extracted from GCNs in both spatial and feature spaces. To solve the boundary confusion in both single-face and multi-face scenarios, the proposed category-aware edge detection module exploits more semantic information than the binary edge detection module used in EARGNet \cite{te2020edge}. 

To address the problem caused by noisy labels in training datasets, we introduce a cyclically learning scheduler inspired by self-training \cite{Yarowsky1995UnsupervisedWS, self_train_theory, Yarowsky1995UnsupervisedWS, Zoph2020RethinkingPA, Yalniz2019BillionscaleSL, Chen2020LeveragingSL,li2020self} to achieve advanced cyclical self-regulation. 
The proposed CSR contains a self-ensemble strategy that can aggregate a set of historical models to obtain a new reliable model and another self-distillation method that exploits the soft labels generated by the aggregated model to guide the successive model learning. Finally, the proposed CSR iteration alternates between these two procedures, correcting the noisy labels during training and promoting the model generalization. The proposed CSR can significantly promote the reliability of the model and labels in a cyclical training scheduler without introducing extra computation costs.

To summarize, our main contributions are as follows:
\vspace{-0.2cm}
\begin{tight_itemize}
\item We propose a decoupled multi-task network including face parsing, binary edge detection, and category edge detection. The face parsing branch introduces a DDGCN without any extra pooling operation to solve the problem of spatial inconsistency, and an additional category edge detection branch is designed to handle the boundary confusion.
\item We introduce a cyclical self-regulation mechanism during training. The iteration alternates between one self-ensemble procedure, boosting model generalization progressively, and another self-distillation processing, regulating noisy labels.
\item Our method establishes new state-of-the-art performance on the Helen\cite{helen} (93.8\% overall F1 score), LaPa\cite{lapa} (92.4\% mean F1) and CelebAMask-HQ\cite{CelebAMask-HQ} (86.1\% mean F1) datasets. Compared to EARGNet\cite{te2020edge}, our method utilizes fewer computation resources as the edge prediction modules can be decoupled from the whole network, decreasing the inference time from 89ms to 31ms but achieving much better performance.
\end{tight_itemize}

\begin{figure*}
\centering
\includegraphics[width=0.98\linewidth]{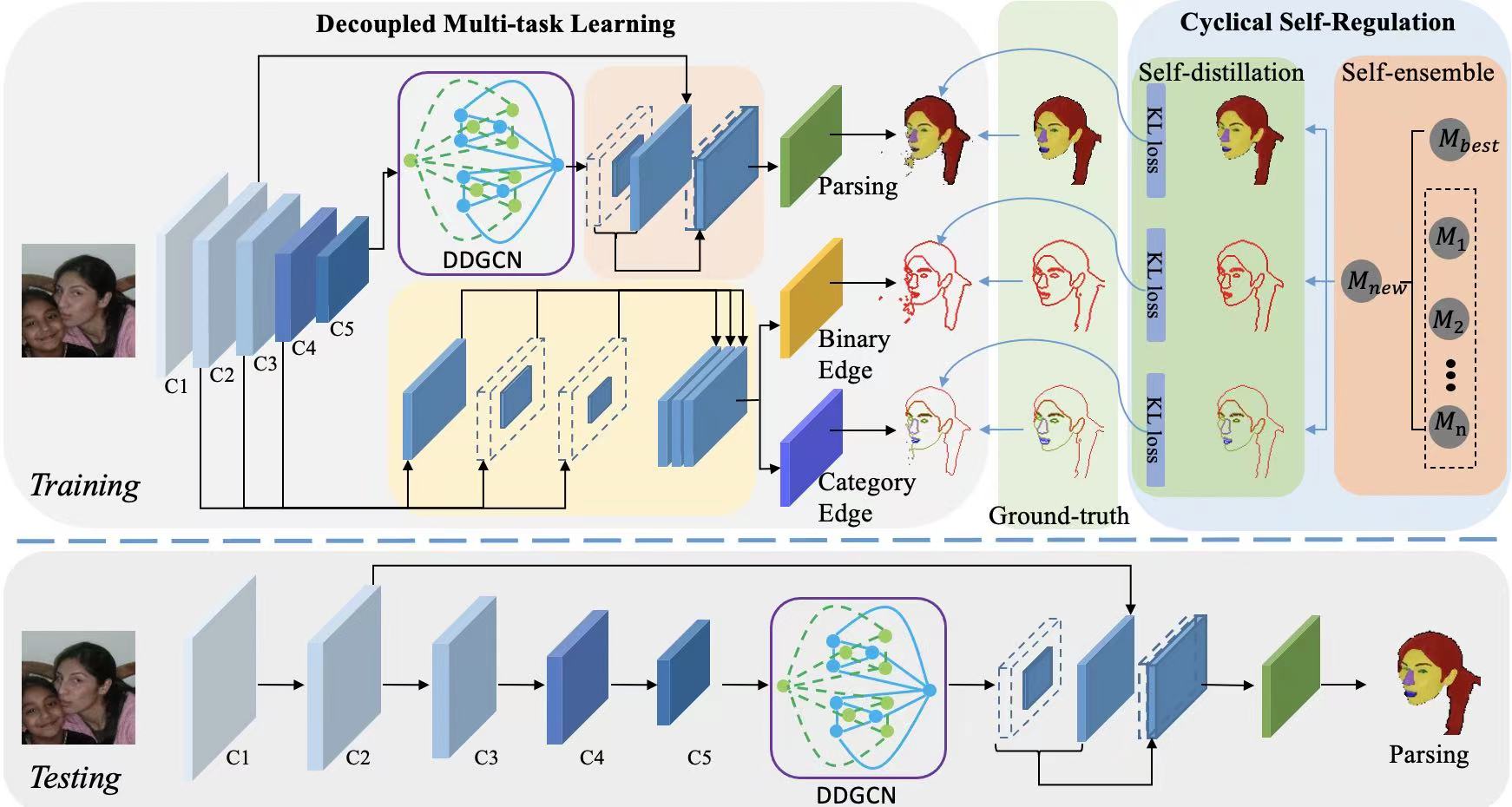}
\vspace{-2.5mm}
\caption{Overview of our proposed DML-CSR method for face parsing. At the training stage, it includes three parallel sub-models of face paring, binary edge detection and category edge detection, jointly trained by a proposed cyclical self-regulation mechanism. At the testing stage, all edge models are decoupled from the whole model.}
\vspace{-6mm}
\label{fig:framwork}
\end{figure*}

\section{Related Work}

\noindent\textbf{Face parsing.} Most existing face parsing methods can be classified into global-based and local-based methods.
Global-based methods aim to predict a pixel-wise label directly from the whole RGB face image. Early works learn spatial correlation between facial parts using various hand-crafted models, such as epitome model \cite{Labelfaces} and exemplar-based method \cite{helen}. Later, many works \cite{multi_objective, cnn_cascade, Zhou2017FacePV, adaptive_field} implant the CNN-based features into the Conditional Random Field (CRF) framework, and adopt a multi-objective learning method to model pixel-wise labels and neighborhood dependencies simultaneously. Lin \etal \cite{roiTanh} design a CNN-based framework with a RoI Tanh-Warping operator to use both central and peripheral information. Te \etal \cite{te2020edge} introduce an edge-aware graph module to effectively reason relationship between facial regions. These global-based approaches inherently integrate the prior into the face layout, but limit accuracy due to overlook on each individual part.

Local-based methods aim to predict each facial part individually by training separated models for different facial regions. Luo \etal \cite{hierarchical_luo} exploit a hierarchical approach to segment each detected facial component separately. Zhou \etal \cite{icnn} propose an interlinked CNN-based model to forecast pixel categories after face detection, taking a large expense of memory and computation consumption. Later, Liu \etal \cite{lius} combines a shallow CNN and a spatially variant RNN in two successive stages to parse a face image at a very fast inference speed. These local-based approaches almost take the coarse-to-fine policy with consideration of both global consistency and local precision. However, it ignores the improvement of accuracy and efficiency from backbone sharing and joint optimization.

\noindent\textbf{Multi-task learning} is a common strategy which jointly trains various tasks through the shared feature mechanism or hidden layers of a ``backbone" model \cite{multi_task}. 
It has been widely applied for solving multiple pixel-level tasks. In the context of deep learning, multi-task learning can be categorized into hard or soft parameter sharing schemes.
In hard parameter sharing based multi-task learning for image segmentation, the parameter set consists of shared and task-specific parameters. UberNet \cite{ubernet} is the first hard parameter sharing model for image segmentation, where a large number of low-, mid-, and high-level image vision tasks are tackled concurrently. Later, most multi-task learning models \cite{Kendall_2018_CVPR, Neven2017FastSU, MultiNet} follow the hard parameter sharing schemes and simply share the same encoder layers. In these works, each task-specific decoding head tails at the end of the shared encoder, leading to sub-optimal task groupings.

In soft parameter sharing based multi-task learning for image segmentation, each task has its own group of parameters, and a feature sharing mechanism is used to handle the cross-task communication. Cross-stitch network \cite{Misra_2016_CVPR} is a typical multi-task architecture adopting the soft-parameter sharing schemes. It linearly combines the activations from every task-specific layer, regarding as soft feature fusion strategy. Afterwards, Ruder \etal \cite{Ruder2019LatentMA} extends this method to learn the selective sharing layers. 
Compared to the hard parameter sharing approaches, the problem of multi-task learning based on soft parameter sharing approaches is a lack of scalability, as the growth of tasks make the size of the multi-task network increase linearly \cite{multi_task_survey}.

\section{Methodology}

This section starts with the analysis of representative failure cases when applying EARGNet \cite{te2020edge} to face parsing. These issues motivate the proposal of a more accurate and robust training method, called Decoupled Multi-task Learning with Cyclical Self-Regulation (DML-CSR). The overall pipeline is illustrated in \figureautorefname{ \ref{fig:framwork}}.

\subsection{Limitations of EAGRNet}

Even though EAGRNet \cite{te2020edge} achieves notable performance on face parsing, it has the following issues during training on public benchmark datasets (\eg Helen \cite{helen}, CelebAMask-HQ \cite{CelebAMask-HQ} and LaPa \cite{lapa}). 

\noindent{\bf Spatial Inconsistency.} 
As shown in the first-row of \figureautorefname{ \ref{fig:failure_case_of_fcn}}, EAGRNet improperly predicts ``neck'' pixels within the cloth area, resulting in spatial inconsistency of cloth. As EAGRNet employs an adaptive average pooling within PSP module \cite{psp} to capture global contextual information, the detailed spatial relationship and constraint between original pixels may be neglected. Therefore, a small part of area within a large region can be predicted as wrong classes. Since directly adopting the general object segmentation method to face parsing is sub-optimal, we explore to avoid the unnecessary pooling operation in our model design.

\noindent{\bf Boundary Confusion.}  As intuitively illustrated in the second-row of \figureautorefname{ \ref{fig:failure_case_of_fcn}}, EARGNet fails to distinguish boundaries between (1) ``cloth'' and ``hair'', and (2) the target face and the surrounding face under crowded scenario. 
Generally, component boundaries between different facial organs and instance boundaries between close faces can be confusing for face parsing models.
As the edge network built in EARGNet simply integrates the binary edge prior into contextual features by the dot product and the pooling operation, it only recovers partial boundaries of regions.

\noindent{\bf Impact from Label Noise.} As the pixel-level annotation is difficult and expensive, most of the face parsing benchmarks (\eg 
Helen \cite{helen} and LaPa \cite{lapa}) are annotated in a semi-automatic approach. Therefore, label noises inevitably exist in these datasets.
As given in the last row of \figureautorefname{ \ref{fig:failure_case_of_fcn}}, annotators mark the ``eyes'' as ``glasses''.
Such annotation errors can limit the model performance, especially for tail classes (\eg ``necklace''). Nevertheless, the EARGNet method is a fully supervised method and lacks a regulation mechanism to tackle label noise.

\subsection{Decoupled Multi-task Learning}

Based on above analysis, we propose an end-to-end decoupled multi-task network to solve problems of spatial inconsistency and boundary confusion. Herein, we define three parallel tasks of face parsing, binary edge detection and category edge detection. 
To prevent using any pooling operation in context embedding, a customized GCN \cite{kipf2017semi} module is designed to gain global contextual relationships for the parsing branch. To alleviate the boundary confusion, a binary edge detection branch as well as a category-aware semantic edge detection branch are jointly trained to gain rich edge information. During training, feature representations are simultaneously optimized for these three tasks, but the auxiliary edge prediction branches are removed during testing, without introducing any extra computation cost. 

An overview of our model architecture is depicted in \figureautorefname{ \ref{fig:framwork}}. Given an input facial image, the ResNet-101 \cite{resnet101} pre-trained on ImageNet is taken as backbone to extract features from different levels, marked as $\{C_1, C_2, C_3, C_4, C_5\}$. Afterwards, remaining parts involve: (1) a face parsing branch, which consists of a context embedding and a parsing head \cite{te2020edge}, (2) a binary edge detection branch utilizing the same edge decoder as \cite{ce2p}, and (3) a category edge detection branch, which features abundant information of component edges. Each task shares same feature representations of first four layers in the backbone model. For the edge detection branches, feature maps from $C_2$, $C_3$ and $C_4$ are concatenated as input. For the parsing branch, context embedding features from $C_5$ are concatenated with the feature maps from $C_2$ as the input.
Since edge branches preserve boundary information in low-level feature maps, joint edge prediction can assist high-level semantic predictions. At the testing phase, these two edge branches are decoupled from the whole model, avoiding extra computation overhead.

\begin{figure}
\centering
\includegraphics[width=\linewidth]{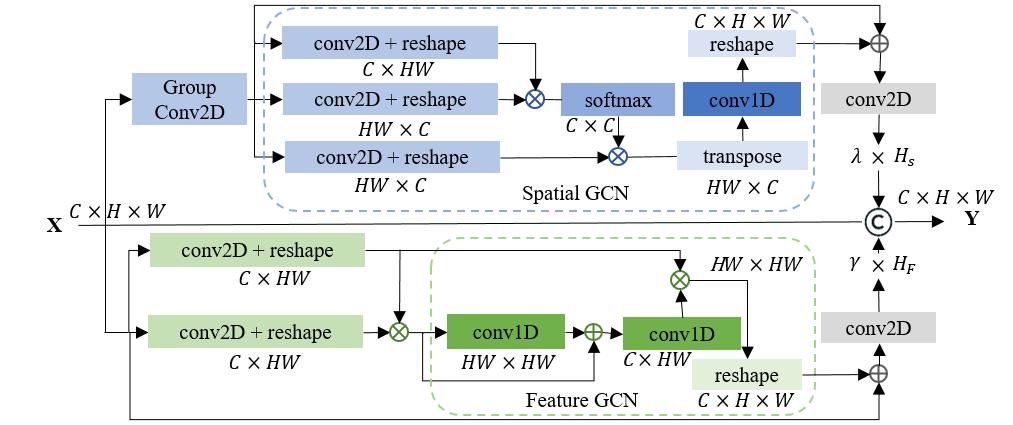}
\vspace{-6mm}
\caption{Illustration of the proposed DDGCN for context embedding. DDGCN is composed of two branches, and each consists of a Graph Convolutional Network (GCN) to model contextual information in the spatial-dimensions and feature-dimensions for a convolutional feature map $X$. No pooling step is involved in DDGCN to avoid spatial inconstancy.}
\label{fig:ddgcn}
\vspace{-4mm}
\end{figure}

\noindent\textbf{Context Embedding without Pooling.} 
Context embedding is crucial for face parsing \cite{psp, NonLocal2018, dilatedConv1, YuKoltun2016}, but the pooling operation results in the problem of spatial inconsistency. To this end, we design a Dynamic Dual Graph Convolution Network (DDGCN), which exploits 1D convolution to build adjacent matrix of GCN over different 2D dimensions. As shown in \figureautorefname{ \ref{fig:ddgcn}}, the proposed DDGCN comprises one weighted GCN (labeled as $H_{S}$) with parameter $\lambda$ in the spatial space and another weighted GCN (labeled as $H_{F}$) with parameter $\gamma$ in the feature space. 
\begin{equation}
\small
Y = X \ \copyright \ (\lambda \times H_{S}) \ \copyright \ (\gamma \times H_{F}),
\end{equation}
where $\copyright$ denotes the operation of concatenation. The parameters $\lambda$ and $\gamma$ are learnable weights for both $H_{S}$ and $H_{F}$, respectively. Different from DGCN \cite{zhangli_dgcn}, we remove the pooling operation during coordinate space projection and, we merge spatial and channel features into the input $X$ via a dynamic concatenation instead of the addition operation. To avoid buffer storage for gradient computation, all BN layers are replaced by Inplace-ABN \cite{inplace}. As the proposed DDGCN is only applied to the $C_5$ feature map, our context embedding is more efficient than EAGRNet, which employs low-level features for graph representation learning.

\begin{figure}
\centering
\subfloat[Binary edge generation]{
\label{fig:bedge}
\includegraphics[height=0.11\textwidth]{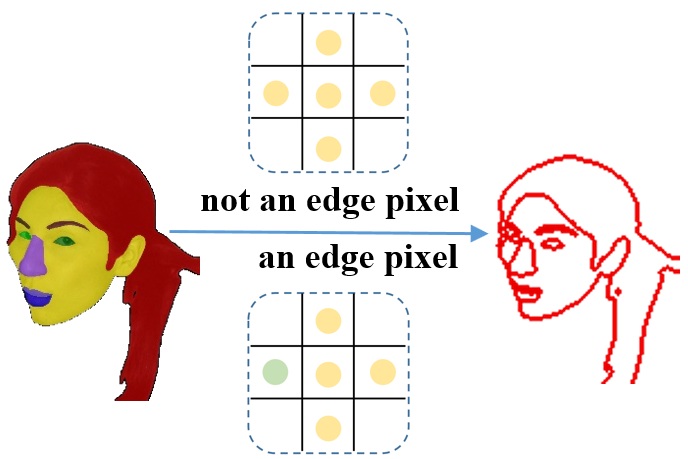}}
\subfloat[Category-aware edge generation]{
\label{fig:cdge}
\includegraphics[height=0.11\textwidth]{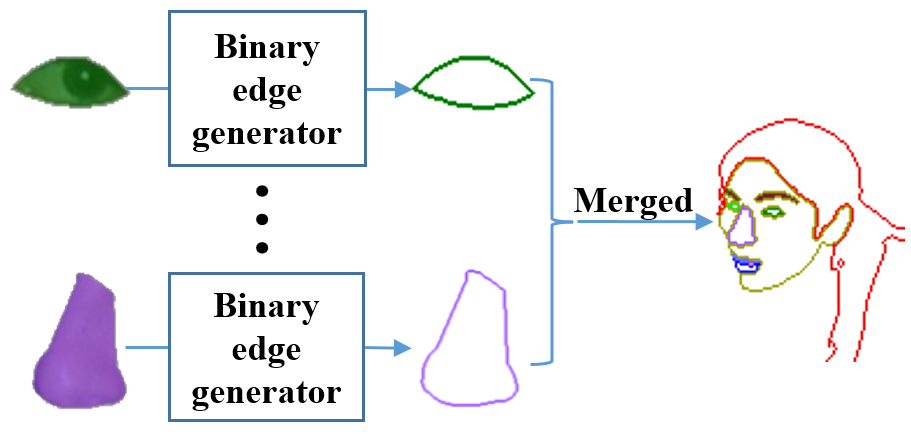}}
\vspace{-2.5mm}
\caption{Binary edge label generation and category-aware edge label generation from the pixel-wise label map.}
\vspace{-4mm}
\label{fig:edgegeneration}
\end{figure}

\noindent\textbf{Binary and Category Edge Assisted Face Parsing.} 
As current training datasets for face parsing do not provide labels for the boundary detection, we first generate pseudo labels of binary and category-aware edges as illustrated in \figureautorefname{ \ref{fig:edgegeneration}}. More specifically, binary edge pixels are identified from the pixel-wise label map by referring the neighboring four pixels. If there exists one neighboring pixel of zero value, the current pixel is regarded as an edge pixel. By employing the same criteria, the category-aware edges are generated independently for each facial component.

To learn shared features for the layers $\{C_1, C_2, C_3, C_4\}$ by simultaneously training the parsing and edge detection tasks, we design a loss function for each task and then sum them together with different weights. Different from the general semantic segmentation, face parsing features on tiny components. To retain the structure of small components, we also employ the Lov{\'a}sz-softmax \cite{lovasz} loss, which utilizes the mean intersection-over-union score to measure difference between ground truth and predicted mask. Hence, the cross-entropy \cite{bce} and Lov{\'a}sz-softmax \cite{lovasz} losses are combined together to optimize the parsing module. Additionally, the weighted cross-entropy \cite{bce} loss is employed to optimize both binary and category-aware edge detection. Consequently, the total multi-task loss is defined as
\small
\begin{equation}
\mathcal{L}_{MT} = \underbrace{\lambda_{0} \cdot (\mathcal{L}^p_{ce} + \mathcal{L}^p_{lov\acute{a}sz})}_{parse} +  \underbrace{\lambda_{1} \cdot \mathcal{L}^{b}_{ce} + \lambda_{2} \cdot \mathcal{L}^{c}_{ce}}_{edges},
\end{equation} 
where $\mathcal{L}^{b}_{ce}$ and $\mathcal{L}^{c}_{ce}$ represent the weighted cross-entropy losses \cite{bce} corresponding to binary and category-aware semantic edges, respectively. The hyper-parameters $\lambda_{0}$, $\lambda_{1}$, and $\lambda_{2}$ denote the different weights for each task.

Besides the above parallel optimization, we also develop a boundary assisted semantic loss which enlarges the parsing loss of boundary pixels according to the binary and category-aware boundary maps. As edge maps are highly related to segmentation maps, it is beneficial to inject two types of edge cues into the parsing module to improve the segmentation accuracy for the components with clear contours. 
To this end, we define a dual edge attention loss
\small
\begin{equation}
\mathcal{L}_{attn}^{b} = \frac{1}{N}\sum_{i=1}^{N} \frac{1}{b_i} * \mathcal{L}_{i}^{p} \odot B_i,
\end{equation}
\begin{equation}
\small
\mathcal{L}_{attn}^{c} = \frac{1}{NC}\sum_{i=1}^{N} \sum_{j=1}^{C}  w_j* \frac{1}{c_{ij}}* \mathcal{L}_{i}^{p} \odot C_{ij},
\label{eq:dal}
\end{equation}
\normalsize
where $N$ is the total number of images in a batch, $b_i$ is the number of boundary pixels in a binary edge label map $B_i \in \mathbb{R}^{H \times W}$, $c_{ij}$ is the number of boundary pixels of a specific category $j$ in a category-aware edge label map $C_{ij} \in \mathbb{R}^{H \times W}$, $w_j$ is a category-aware weight to emphasize a  specific class $j$ (\eg the tail class of ``necklace') which can increase the weights of tail classes, and $\mathcal{L}_{i}^{p} \in \mathbb{R}^{H \times W} $ is the cross-entropy between a predicted parsing result and the ground-truth. Different from the binary boundary attention loss proposed in \cite{lapa}, we further introduce category-aware boundary-attention semantic loss, significantly improving segmentation results of underrepresented classes.

The overall loss of our decoupled multi-task learning can be summarized as
\small
\begin{equation}
\begin{aligned}
\mathcal{L}_{DML} &= \underbrace{\lambda_{0} \cdot (\mathcal{L}^p_{ce} + \mathcal{L}^p_{lov\acute{a}sz})}_{parse} \\
&+  \underbrace{\lambda_{1} \cdot \mathcal{L}^{b}_{ce} + \lambda_{3} \cdot \mathcal{L}_{attn}^{b}}_{binary-edge} 
+ \underbrace{\lambda_{2} \cdot \mathcal{L}^{c}_{ce} + \lambda_{4} \cdot \mathcal{L}_{attn}^{c}}_{category-edge},
\end{aligned}
\label{eq:dmlloss}
\end{equation}
\normalsize
where $\lambda_{3}$ and $\lambda_{4}$ correspond to weights of attention losses for binary and category-aware edges, respectively.

\subsection{Cyclical Self-Regulation}

To alleviate label noise, we introduce a Cyclical Self-regulation (CSR) training strategy to achieve online refinement labels. The proposed CSR depicted in \figureautorefname{ \ref{fig:framwork}} includes two parts, self-ensemble  and self-distillation.
 
\noindent\textbf{Model Generalization via Self-Ensemble.}  As illustrated in the self-ensemble process of \figureautorefname{ \ref{fig:framwork}}, given a best model $M_{best}$ from previous epochs and a set of next successive models \{$M_1, M_2, \dots, M_n$\}, a new model is obtained by aggregating weights of these models
\begin{equation}
\small
M = \frac{k}{k + 1} M_{best} + \frac{1}{(k+1)N} \sum_{n=1}^{N}M_{n},
\end{equation}
where $k$ is the current cycle number and $1\leq k\leq K$, and $n$ is the number of models used in a cycle and $1\leq n\leq N$. Moreover, symbols $M$, $M_{best}$ and $M_{n}$ represent the weights of aggregated, best and current models, respectively. In addition, all training data is forwarded into new aggregated model to re-estimate the statistical parameters in all Inplace-ABN \cite{inplace} layers.   

\noindent\textbf{Label Refinement via Self-Distillation.} 
As the soft labels contain dark knowledge \cite{Hinton2015DistillingTK} and less label noise, we explore self-distillation to improve the parsing performance. More specifically, as shown in the self-distillation process of \figureautorefname{ \ref{fig:framwork}}, the parsing results generated from the above aggregated model are exploited to supervise the multi-task learning. The total weighted loss is defined as  
\begin{equation}
\small
\mathcal{L}_{CSR} = \underbrace{\alpha_{0} \cdot (\mathcal{L}^p_{kl} + \mathcal{L}^p_{lov\acute{a}sz})}_{parse} +  \underbrace{\alpha_{1} \cdot \mathcal{L}_{kl}^{b} + \alpha_{2} \cdot \mathcal{L}_{kl}^{c}}_{edges},
\end{equation} 
where $\mathcal{L}_{kl}^p$, $\mathcal{L}_{kl}^{b}$, $\mathcal{L}_{kl}^{c}$ represent the Kullback-Leibler divergence losses \cite{bce} for face parsing, binary edge and category-aware edge tasks, respectively. They compute the difference between soft labels of the aggregated model and prediction results of the current model. Hyper-parameters $\alpha_{0}$, $\alpha_{1}$, $\alpha_{2}$ are weights assigned to each task.

Finally, both self-ensemble and self-distillation processes mutually iterates in a cycle manner, promoting model generalization and correcting noisy labels progressively.

\section{Experiments}

\begin{table*}
\small
\centering
\begin{tabular}{@{}l|llllllll|lc@{}}
\toprule
Method & Skin & Nose & U-lip & I-mouth & L-lip & Eyes & Brows & Mouth & Overall F1 \\
\midrule
Liu \etal \cite{lius} & \makecell[c]{92.1} & \makecell[c]{93.0} & \makecell[c]{74.3} & \makecell[c]{79.2} & \makecell[c]{81.7} & \makecell[c]{86.8} & \makecell[c]{77.0} & \makecell[c]{89.1} & \makecell[c]{88.6} \\

Guo \etal \cite{guo2018} & \makecell[c]{93.8} & \makecell[c]{94.1} & \makecell[c]{75.8} & \makecell[c]{83.7} & \makecell[c]{83.1} & \makecell[c]{80.4} & \makecell[c]{87.1} & \makecell[c]{92.4} & \makecell[c]{90.5} \\

Lin \etal \cite{roiTanh} & \makecell[c]{94.5} & \makecell[c]{95.6} & \makecell[c]{79.6} & \makecell[c]{86.7} & \makecell[c]{89.8} & \makecell[c]{89.6} & \makecell[c]{83.1} & \makecell[c]{95.0} & \makecell[c]{92.4} \\

Wei \etal \cite{wei_tip} & \makecell[c]{95.6} & \makecell[c]{95.2} & \makecell[c]{80.0} & \makecell[c]{86.7} & \makecell[c]{86.4} & \makecell[c]{89.0} & \makecell[c]{82.6} & \makecell[c]{93.6} & \makecell[c]{91.6} \\

Liu \etal \cite{lapa} & \makecell[c]{94.9} & \makecell[c]{95.8} & \makecell[c]{83.7} & \makecell[c]{89.1} & \makecell[c]{\textbf{91.4}} & \makecell[c]{89.8} & \makecell[c]{83.5} & \makecell[c]{\textbf{96.1}} & \makecell[c]{93.1} \\

Te \etal \cite{te2020edge} &\makecell[c]{94.6} & \makecell[c]{\textbf{96.1}} & \makecell[c]{83.6} & \makecell[c]{89.8} & \makecell[c]{91.0} & \makecell[c]{90.2} & \makecell[c]{84.9} & \makecell[c]{95.5} & \makecell[c]{93.2} \\
\midrule
DML-CSR (Ours) & \makecell[c]{\textbf{96.6}} & \makecell[c]{95.5} & \makecell[c]{\textbf{87.6}} & \makecell[c]{\textbf{91.2}} & \makecell[c]{91.2} & \makecell[c]{\textbf{90.9}} & \makecell[c]{\textbf{88.5}} & \makecell[c]{95.9} & \makecell[c]{\textbf{93.8}} \\
\bottomrule
\end{tabular}
\vspace{-2.5mm}
\caption{Comparison with state-of-the-art methods on the Helen dataset in overall F1 score.}
\vspace{-2.5mm}
\label{tab:comparision_helen}
\end{table*}

\begin{table*}
\small
\centering
\begin{tabular}{@{}l|llllllllll|c@{}}
\toprule
Method & Skin & Hair & L-Eye & R-Eye & U-lip & I-mouth & L-lip & Nose & L-Brow & R-Brow & Mean F1 \\
\midrule
Zhao \etal \cite{psp} & \makecell[c]{93.5} & \makecell[c]{94.1} & \makecell[c]{86.3} & \makecell[c]{86.0} & \makecell[c]{83.6} & \makecell[c]{86.9} & \makecell[c]{84.7} & \makecell[c]{94.8} & \makecell[c]{86.8} & \makecell[c]{86.9} & \makecell[c]{88.4} \\

Liu \etal \cite{lapa} & \makecell[c]{97.2} & \makecell[c]{96.3} & \makecell[c]{88.1} & \makecell[c]{88.0} & \makecell[c]{84.4} & \makecell[c]{87.6} & \makecell[c]{85.7} & \makecell[c]{95.5} & \makecell[c]{87.7} & \makecell[c]{87.6} & \makecell[c]{89.8} \\

Te \etal \cite{te2020edge} & \makecell[c]{97.3} & \makecell[c]{96.2} & \makecell[c]{89.5} & \makecell[c]{90.0} & \makecell[c]{\textbf{88.1}} & \makecell[c]{90.0} & \makecell[c]{89.0} & \makecell[c]{97.1} & \makecell[c]{86.5} & \makecell[c]{87.0} & \makecell[c]{91.1} \\

\midrule
DML-CSR (Ours) & \makecell[c]{\textbf{97.6}} & \makecell[c]{\textbf{96.4}} & \makecell[c]{\textbf{91.8}} & \makecell[c]{\textbf{91.5}} & \makecell[c]{88.0} & \makecell[c]{\textbf{90.5}} & \makecell[c]{\textbf{89.9}} & \makecell[c]{\textbf{97.3}} & \makecell[c]{\textbf{90.4}} & \makecell[c]{\textbf{90.4}} & \makecell[c]{\textbf{92.4}} \\
\bottomrule
\end{tabular}
\vspace{-2.5mm}
\caption{Comparison with state-of-the-art methods on the LaPa dataset in mean F1. }
\vspace{-2.5mm}
\label{tab:comparision_lapa}
\end{table*}

\begin{table*}
  \small
  \centering
  \begin{tabular}{@{}l|lllllllll|c@{}}
    \toprule
    \multirow{2}{*}{Method} & Face & Nose & Glasses & L-Eye & R-Eye & L-Brow & R-Brow & L-Ear & R-Ear & \multirow{2}{*}{Mean F1} \\
     & I-Mouth & U-Lip & L-Lip & Hair & Hat & Earring & Necklace & Neck & Cloth \\
    \midrule
    \multirow{2}{*}{Zhao \etal \cite{psp}} & \makecell[c]{94.8} & \makecell[c]{90.3} & \makecell[c]{75.8} & \makecell[c]{79.9} & \makecell[c]{80.1} & \makecell[c]{77.3} & \makecell[c]{78.0} & \makecell[c]{75.6} & \makecell[c]{73.1} & \multirow{2}{*}{76.2}\\ 
                                & \makecell[c]{89.8} & \makecell[c]{87.1} & \makecell[c]{88.8} & \makecell[c]{90.4} & \makecell[c]{58.2} & \makecell[c]{65.7} & \makecell[c]{19.4} & \makecell[c]{82.7} & \makecell[c]{64.2} \\
    \hline
    \multirow{2}{*}{Lee \etal \cite{CelebAMask-HQ}} & \makecell[c]{95.5} & \makecell[c]{85.6} & \makecell[c]{\textbf{92.9}} & \makecell[c]{84.3} & \makecell[c]{85.2} & \makecell[c]{81.4} & \makecell[c]{81.2} & \makecell[c]{84.9} & \makecell[c]{83.1} & \multirow{2}{*}{80.3}\\ 
                                & \makecell[c]{63.4} & \makecell[c]{88.9} & \makecell[c]{90.1} & \makecell[c]{86.6} & \makecell[c]{\textbf{91.3}} & \makecell[c]{63.2} & \makecell[c]{26.1} & \makecell[c]{\textbf{92.8}} & \makecell[c]{68.3} \\
    \hline
    \multirow{2}{*}{Luo \etal \cite{ehanet}} & \makecell[c]{96.0} & \makecell[c]{93.7} & \makecell[c]{90.6} & \makecell[c]{86.2} & \makecell[c]{86.5} & \makecell[c]{83.2} & \makecell[c]{83.1} & \makecell[c]{86.5} & \makecell[c]{84.1} & \multirow{2}{*}{84.0}\\ 
                                & \makecell[c]{93.8} & \makecell[c]{88.6} & \makecell[c]{90.3} & \makecell[c]{93.9} & \makecell[c]{85.9} & \makecell[c]{67.8} & \makecell[c]{30.1} & \makecell[c]{88.8} & \makecell[c]{83.5} \\
    \hline
    \multirow{2}{*}{Te \etal \cite{te2020edge}} & \makecell[c]{\textbf{96.2}} & \makecell[c]{\textbf{94.0}} & \makecell[c]{92.3} & \makecell[c]{88.6} & \makecell[c]{88.7} & \makecell[c]{\textbf{85.7}} & \makecell[c]{85.2} & \makecell[c]{88.0} & \makecell[c]{85.7} & \multirow{2}{*}{85.1}\\ 
                            & \makecell[c]{\textbf{95.0}} & \makecell[c]{\textbf{88.9}} & \makecell[c]{\textbf{91.2}} & \makecell[c]{\textbf{94.9}} & \makecell[c]{87.6} & \makecell[c]{68.3} & \makecell[c]{27.6} & \makecell[c]{89.4} & \makecell[c]{85.3} \\
    \midrule
    \multirow{2}{*}{DML-CSR (Ours)} & \makecell[c]{95.7} & \makecell[c]{93.9} & \makecell[c]{92.6} & \makecell[c]{\textbf{89.4}} & \makecell[c]{\textbf{89.6}} & \makecell[c]{85.5} & \makecell[c]{\textbf{85.7}} & \makecell[c]{\textbf{88.3}} & \makecell[c]{\textbf{88.2}} & \multirow{2}{*}{ \textbf{86.1}}\\ 
                                & \makecell[c]{91.8} & \makecell[c]{87.4} & \makecell[c]{91.0} & \makecell[c]{94.5} & \makecell[c]{88.5} & \makecell[c]{\underline{\textbf{71.4}}} & \makecell[c]{\underline{\textbf{40.6}}} & \makecell[c]{89.6} & \makecell[c]{\textbf{85.7}} \\
    \bottomrule
  \end{tabular}
  \vspace{-2.5mm}
  \caption{Comparison with state-of-the-art methods on the CelebAMask-HQ dataset in mean F1. }
  \vspace{-2.5mm}
  \label{tab:comparision_celebA} 
\end{table*}

\begin{table*}
  \small
  \centering
  \resizebox{\linewidth}{!}{
  \begin{tabular}[width=\linewidth]{ cccc|cc|cc|cc } 
    \hline
    \multirow{2}{3em}{Baseline} & \multirow{2}{3em}{DDGCN} & \multirow{2}{2em}{DML} & \multirow{2}{2em}{CSR} & \multicolumn{2}{c|}{Helen} & \multicolumn{2}{c|}{CelebAMask-HQ} & \multicolumn{2}{c}{LaPa}  \\
    &&&& Mean IoU & Overall F1 & Mean IoU & Mean F1 & Mean IoU & Mean F1   \\
    \hline \hline
     \checkmark  &&&&  82.36 & 92.11 & 76.14 &  84.34 & 83.16 & 89.84 \\
     \checkmark  & \checkmark &&&  83.42 {\scriptsize (\textcolor{color2}{+ 1.06})} & 92.56 {\scriptsize (\textcolor{color2}{+ 0.45})} & 77.41 {\scriptsize (\textcolor{color2}{+ 1.27})} & 85.33 {\scriptsize (\textcolor{color2}{+ 0.99})} & 86.65 {\scriptsize (\textcolor{color2}{+ 3.49})} & 92.10 {\scriptsize (\textcolor{color2}{+ 2.26})} \\
     \checkmark  & \checkmark & \checkmark && 85.48 {\scriptsize (\textcolor{color2}{+ 3.12})} & 93.75 {\scriptsize (\textcolor{color2}{+ 1.64})} & 77.69 {\scriptsize (\textcolor{color2}{+ 1.55})} & 85.98 {\scriptsize (\textcolor{color2}{+ 1.64})} & 87.00 {\scriptsize (\textcolor{color2}{+ 3.84})} & 92.32 {\scriptsize (\textcolor{color2}{+ 2.48})} \\
     \checkmark  & \checkmark & \checkmark & \checkmark &  \textbf{85.58} {\scriptsize (\textcolor{color2}{+ 3.22})} & \textbf{93.78} {\scriptsize (\textcolor{color2}{+ 1.67})} &  \textbf{77.81} {\scriptsize (\textcolor{color2}{+ 1.67})} & \textbf{86.07} {\scriptsize (\textcolor{color2}{+ 1.73})} & \textbf{87.13} {\scriptsize (\textcolor{color2}{+ 3.97})} & \textbf{92.38} {\scriptsize (\textcolor{color2}{+ 2.54})} \\
    \hline
  \end{tabular}}
  \vspace{-2.5mm}
  \caption{Ablation study of DML-CSR on the Helen, CelebAMask-HQ and LaPa datasets. Here, DDGCN is used for context embedding. DML denotes the multi-task learning for our decoupled model including face parsing, binary and category edge detection. CSR represents the cyclical self-regulation.}
  \vspace{-5mm}
  \label{tab:ablation}
\end{table*}

\noindent{\bf Datasets.}
We use Helen \cite{helen}, CelebAMask-HQ \cite{CelebAMask-HQ}, and LaPa \cite{lapa} for experiments. The Helen dataset contains 2,330 images with 11 labels: ``background", ``facial skin", ``left/right brow", ``left/right eye", ``nose", ``upper/lower lip", ``inner mouth" and ``hair". It is split into 2,000, 230 and 100 images for training, validation and testing. The CelebAMask-HQ dataset includes 24,183, 2,993, and 2,824  images for training, validation and testing. Apart from the 11 categories of the Helen dataset, the CelebAMask-HQ dataset adds extra 8 classes, including ``left/right ear', ``eyeglass", ``earing", ``necklace", ``neck" and ``cloth". The LaPa dataset features rich variations in expression, pose and occlusion, consisting of 11 categories as the Helen dataset. It is partitioned into 18,176 samples for training, 2,000 samples for validation, and 2,000 samples for testing.

\noindent{\bf Implementation Details.}
The proposed method is implemented by Pytorch \cite{pytorch}, adopting the ResNet101 \cite{resnet101} as a backbone. The weights of the backbone are initialized with the pre-trained model on ImageNet \cite{imagenet}. Batch normalizations in our network are all replaced by In-Place Activated Batch Norm \cite{inplace}. The input image size is $473\times473$ at both training and testing stages. During training, the data is augmented using: random rotation selecting an angle within (-30\textdegree, 30\textdegree) and random scaling with a factor from 0.75 to 1.25. We set the batch size as 28 and the network is trained for 200 epochs in total. The first 150 epochs are trained as initialization, following $K=5$ cycles and each containing $N=10$ epochs of the self-training process. 

During the decoupled multi-task learning, we follow the similar training strategies as EAGRNet \cite{te2020edge}, \ie Stochastic Gradient Descent (SGD) optimizer with the base learning rate 0.001 and the weight of decay of 0.0005. For the total loss function, weights of parsing, binary edge and category edge losses are set as $\lambda_{0} = 1$, $\lambda_{1} = 1$, and $\lambda_{2} = 1$. respectively. To recover boundaries of tail classes (\eg necklace and earring), weights $\lambda_{3} = 4$ and $\lambda_{4} = 1$ are assigned to both binary and category edge attention losses, respectively. For the cyclical self-regulation, the cosine annealing learning rate scheduler \cite{li2020self} with a learning rate of $10^{-5}$ is employed to optimize the model generalization. The weights of self-distillation losses for parsing, binary and category-aware edges are set to $\alpha_{0} = 1$, $\alpha_{1} = 1$ and $\alpha_{2} = 0.1$. 

\noindent{\bf Evaluation Metrics.}
To measure the performance of a face parsing model, two universally accepted evaluation metrics are employed, namely mean Intersection over Union (mIoU) and F1 score. To keep consistent comparison with the previous methods, the overall F1-score on the Helen dataset is calculated over the merged facial components: brows (left and right), eyes (left and right), nose,  and mouth (upper lip, lower lip, and inner mouth). For the CelebAMask-HQ and LaPa datasets, the mean F1-score is computed over all categories excluding the background.

\begin{table}
  \small
  \centering
\resizebox{\linewidth}{!}{
  \begin{tabular}[width=\linewidth]{ l|c|c|c } 
    \hline 
    \multirow{2}{4em}{Method} & {\footnotesize Helen} & {\footnotesize CelebAMask-HQ} & {\footnotesize LaPa}  \\
& {\footnotesize Overall F1} & {\footnotesize Mean F1} & {\footnotesize Mean F1}  \\
\hline\hline
Baseline & 92.11 & 84.34 & 89.84 \\
\hline
+PSP \cite{psp} & 92.20 & 84.76 & 90.80\\
+PSP-pooling & 92.37  &  84.83 & 91.35  \\
+DGCNet \cite{zhangli_dgcn} & 92.41 & 85.17 & 91.72 \\ 
+DGCNet-pooling &  92.45 &  85.20 &  91.99 \\
\hline
+DDGCN  & \textbf{92.56} & \textbf{85.33} & \textbf{92.10} \\ 
\hline
\end{tabular}}
\vspace{-2.5mm}
\caption{Comparisons of different contextual modules on the parsing branch. Here, ``+'' means that the context embedding is added into the baseline, and ``-pooling'' denotes that the pooling operation is removed from the context embedding.}
\vspace{-2mm}
\label{tab:ddgcn}
\end{table}

\begin{table}
  \small
  \centering
  \begin{tabular}[width=\linewidth]{ l|c|c|c } 
    \hline 
    \multirow{2}{4em}{Method} & {\footnotesize Helen} & {\footnotesize CelebAMask-HQ} & {\footnotesize LaPa}  \\
& {\footnotesize Overall F1} & {\footnotesize Mean F1} & {\footnotesize Mean F1}  \\
    \hline \hline
Baseline & 92.11 & 84.34 & 89.84 \\\hline
+DML$_{p+b}$  & 93.35 & 85.58 & 92.16 \\
+DML$_{p+b+ba}$ & 93.52 & 85.69  & 92.24   \\
+DML$_{p+c}$ &  93.61 &  85.73 &  92.21\\ 
+DML$_{p+c+ca}$  & 93.71 & 85.87 & 92.28 \\ 
+DML$_{p+b+c}$  & 93.65 & 85.80 & 92.26 \\ 
\hline
+DML$_{all}$  & \textbf{93.75} & \textbf{85.98} & \textbf{92.32} \\ 
\hline
\end{tabular}
\vspace{-2.5mm}
\caption{Results of our proposed multi-task learning on the Helen, CelebAMask-HQ and LaPa datasets. Here, ``+" denotes adding multi-task branches into the baseline where DDGCN is used as context embedding. Losses of face parsing, binary edge detection, and category edge detection are denoted as $*_p$, $*_b$ and $*_c$ in the subscript. Binary edge attention and category edge attention losses are denoted as $*_{ba}$ and $*_{ca}$ in the subscript.}
\label{tab:multi_task}
\vspace{-5mm}
\end{table}

\subsection{Comparison with State-of-the-art}
In this paper, we thoroughly compare the performance of our proposed model with existing state-of-the-art methods (\ie Zhao \etal \cite{psp}, Liu \etal \cite{lapa}, Lee \etal \cite{CelebAMask-HQ}, Luo \etal \cite{ehanet}, Liu \etal \cite{lius}, Guo \etal \cite{guo2018}, Lin \etal \cite{roiTanh}, Wei \etal \cite{wei_tip}, and Te \etal \cite{te2020edge}) on the Helen, LaPa and CelebAMask-HQ datasets. 
Statistical results in \tableautorefname { \ref{tab:comparision_helen}},  \tableautorefname { \ref{tab:comparision_lapa}}, and \tableautorefname { \ref{tab:comparision_celebA}} demonstrate that the proposed DML-CSR significantly outperforms other methods, achieving $93.8\%$, $92.4\%$, and $86.1\%$ F1 scores on Helen, LaPa and CelebAMask-HQ, respectively. 
On the Lapa dataset, DML-CSR exhibits obvious advantages on eyebrow parsing. 
On the CelebAMask-HQ dataset, DML-CSR achieves much better performance on tail classes, such as ``earring" and ``necklace". 
Compared to EAGRNet \cite{te2020edge}, DML-CSR reduces the parameters from 66.72M to 59.67M, and decreases the FLOP count from 51.63G to 48.54G. Given an image of the same input size as EAGRNet \cite{te2020edge}, DML-CSR dramatically shortens the inference time from 89ms to 31ms per image. In a word, DML-CSR utilizes fewer computation resources to outperform the state-of-the-art method.

\subsection{Ablation Study}

\noindent\textbf{Analysis of Improvement.} To illustrate the effect of individual modules and training strategy, the model after removing some components is trained from scratch under the same setting. The baseline method adopts the parsing module with a simple convolution unit, which includes a $3\times3$ convolution and an Inplace-ABN \cite{inplace} to map features from the last layer of the backbone into new features of $256$ dimensions. As shown in \tableautorefname{ \ref{tab:ablation}}, our proposed DML-CSR substantially improves performance on face parsing. Compared to our baseline, adopting the DDGCN without any pooling operation as context embedding achieves a significant performance improvement.
Then, appending semantic edge modules to enhance shared features has a further advance on parsing performance.
The best results are obtained by training the decoupled multi-task network in a self-regulation mechanism, resulting in around 3.2\% and 4.0\% improvements of mean IoU on the Helen and LaPa datasets, respectively. Besides, it outperforms the baseline by around 1.7\% overall F1 score improvement on the Helen dataset, and by over $2.5\%$ mean F1 improvement on the LaPa dataset. On the CelebAMask-HQ dataset, DML-CSR also outperforms the baseline by around 1.7\% improvement in both mean IoU and mean F1.  

\noindent{\bf Comparison of Various Contextual Modules.} To prove the effectiveness of our proposed DDGCN for learning contextual representation, the above-mentioned simple convolution unit in the baseline is substituted by various context embedding modules. Ablation experiments in \tableautorefname{ \ref{tab:ddgcn}} show that the pooling operation in PSP \cite{psp} and DGCNet \cite{zhangli_dgcn} is harmful for the performance and the proposed DDGCN surpasses other contextual modules by dropping the pooling step and adopting dynamic feature fusion strategies.
\begin{figure}
\centering
\includegraphics[width=0.7\linewidth]{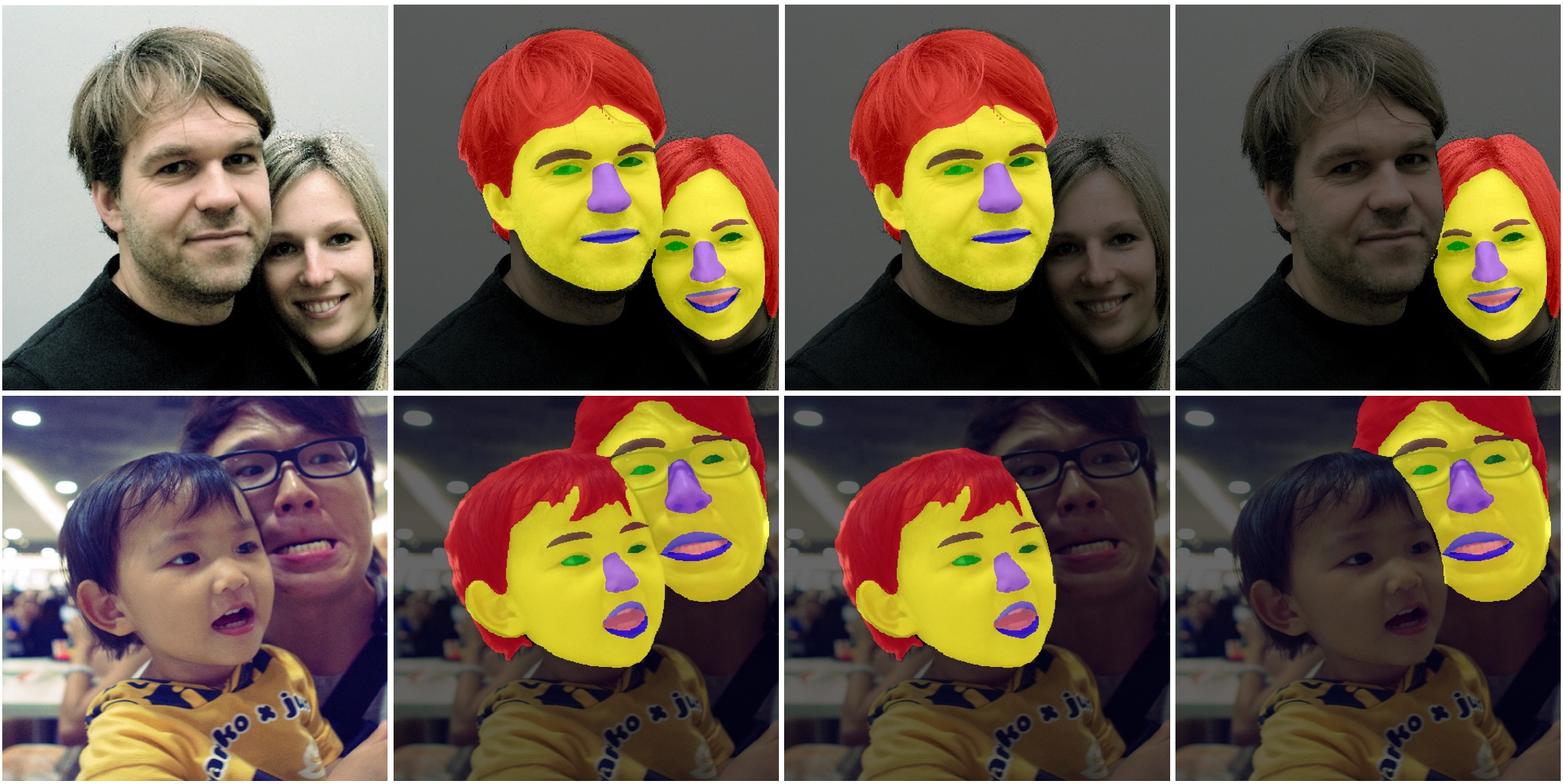}
\vspace{-2.5mm}
\caption{DML-CSR can easily distinguish different face instances under crowded scenarios due to the auxiliary category edge prediction. LaPa model is used here for visualization.}
\label{fig:multiface}
\vspace{-6mm}
\end{figure}

\noindent{\bf Comparison of Different Auxiliary Tasks.}
Visual examples of \figureautorefname{ \ref{fig:multiface}} show that auxiliary category edge modules can distinguish boundaries between facial components and different faces. To further explore the effect of category edge detection, several related experiments are executed.
As we can see from the results listed in \tableautorefname{ \ref{tab:multi_task}}, both the binary edge detection branch and the category-aware edge detection branch can obviously improve the performance of face parsing. However, the category-aware edge is more informative than binary edge, thus it is more beneficial for face parsing. Besides, our proposed dual edge attention loss on the equation \eqref{eq:dal} further improves overall performance of face parsing on three benchmark testing datasets.

\begin{figure}
\subfloat[Image]{
    \begin{minipage}[t]{0.14\linewidth}
    \includegraphics[width=1.3cm, height=1.3cm]{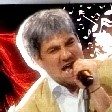}
   \\ \\[-2.5ex]
    \includegraphics[width=1.3cm, height=1.3cm]{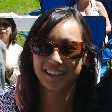}
   \\ \\[-2.5ex]
    \includegraphics[width=1.3cm, height=1.3cm]{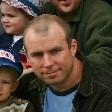}
   \\ \\[-2.5ex]
    \includegraphics[width=1.3cm, height=1.3cm]{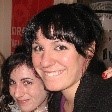}
   \\ \\[-2.5ex]
    \includegraphics[width=1.3cm, height=1.3cm]{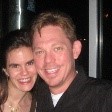}
   \\ \\[-2.5ex]
    \includegraphics[width=1.3cm, height=1.3cm]{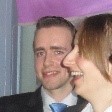}
    \end{minipage}
}
\subfloat[Baseline]{
    \begin{minipage}[t]{0.14\linewidth}
    \includegraphics[width=1.3cm, height=1.3cm]{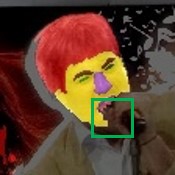}
   \\ \\[-2.5ex]
    \includegraphics[width=1.3cm, height=1.3cm]{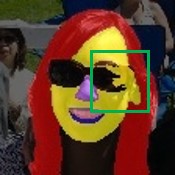}
   \\ \\[-2.5ex]
    \includegraphics[width=1.3cm, height=1.3cm]{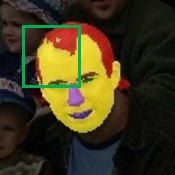}
   \\ \\[-2.5ex]
    \includegraphics[width=1.3cm, height=1.3cm]{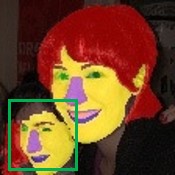}
   \\ \\[-2.5ex]
    \includegraphics[width=1.3cm, height=1.3cm]{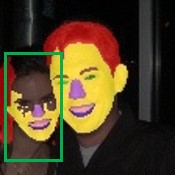}
   \\ \\[-2.5ex]
    \includegraphics[width=1.3cm, height=1.3cm]{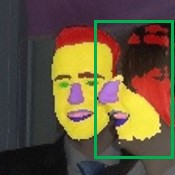}
    \end{minipage}
}
\subfloat[\scriptsize+DDGCN]{
    \begin{minipage}[t]{0.14\linewidth}
    \includegraphics[width=1.3cm, height=1.3cm]{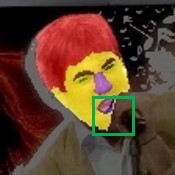}
   \\ \\[-2.5ex]
    \includegraphics[width=1.3cm, height=1.3cm]{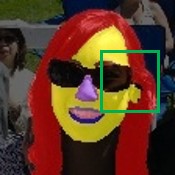}
   \\ \\[-2.5ex]
    \includegraphics[width=1.3cm, height=1.3cm]{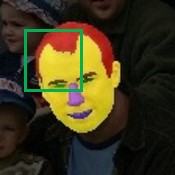}
   \\ \\[-2.5ex]
    \includegraphics[width=1.3cm, height=1.3cm]{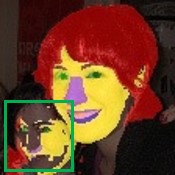}
   \\ \\[-2.5ex]
    \includegraphics[width=1.3cm, height=1.3cm]{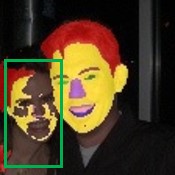}
   \\ \\[-2.5ex]
    \includegraphics[width=1.3cm, height=1.3cm]{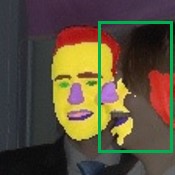}
    \end{minipage}
}
\subfloat[\scriptsize+DML]{
    \begin{minipage}[t]{0.14\linewidth}
    \includegraphics[width=1.3cm, height=1.3cm]{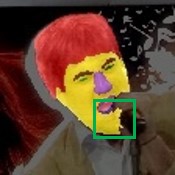}
   \\ \\[-2.5ex]
    \includegraphics[width=1.3cm, height=1.3cm]{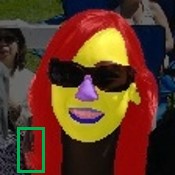}
   \\ \\[-2.5ex]
    \includegraphics[width=1.3cm, height=1.3cm]{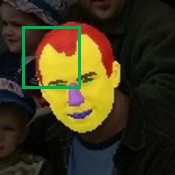}
   \\ \\[-2.5ex]
    \includegraphics[width=1.3cm, height=1.3cm]{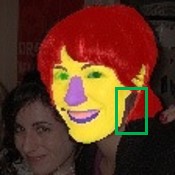}
   \\ \\[-2.5ex]
    \includegraphics[width=1.3cm, height=1.3cm]{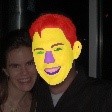}
   \\ \\[-2.5ex]
    \includegraphics[width=1.3cm, height=1.3cm]{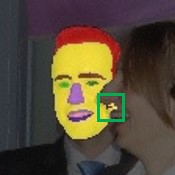}
    \end{minipage}
}
\subfloat[\scriptsize+CSR]{
    \begin{minipage}[t]{0.14\linewidth}
    \includegraphics[width=1.3cm, height=1.3cm]{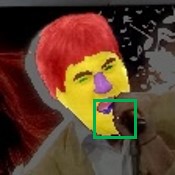}
   \\ \\[-2.5ex]
    \includegraphics[width=1.3cm, height=1.3cm]{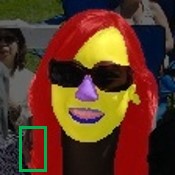}
   \\ \\[-2.5ex]
    \includegraphics[width=1.3cm, height=1.3cm]{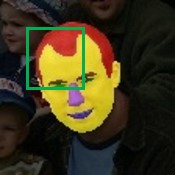}
   \\ \\[-2.5ex]
    \includegraphics[width=1.3cm, height=1.3cm]{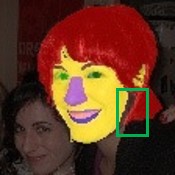}
   \\ \\[-2.5ex]
    \includegraphics[width=1.3cm, height=1.3cm]{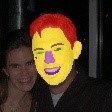}
   \\ \\[-2.5ex]
    \includegraphics[width=1.3cm, height=1.3cm]{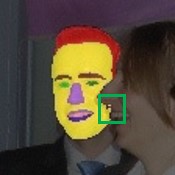}
    \end{minipage}
}
\subfloat[GT]{
    \begin{minipage}[t]{0.13\linewidth}
    \includegraphics[width=1.3cm, height=1.3cm]{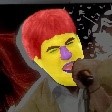}
   \\ \\[-2.5ex]
    \includegraphics[width=1.3cm, height=1.3cm]{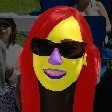}
   \\ \\[-2.5ex]
    \includegraphics[width=1.3cm, height=1.3cm]{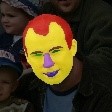}
   \\ \\[-2.5ex]
    \includegraphics[width=1.3cm, height=1.3cm]{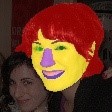}
   \\ \\[-2.5ex]
    \includegraphics[width=1.3cm, height=1.3cm]{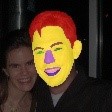}
   \\ \\[-2.5ex]
    \includegraphics[width=1.3cm, height=1.3cm]{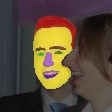}
    \end{minipage}
}
\vspace{-2.5mm}
\caption{DML-CSR can obtain complete facial components with clear boundaries in both single-face and multi-face scenarios. Visual examples in different columns are generated by the corresponding LaPa models. Here, ``+" denotes that the current component is added into the model in the previous column.}
\vspace{-6mm}
\label{fig:visualization}
\end{figure}

\noindent\textbf{Analysis of Visual Results.} To better understand the effect of the proposed methods step-by-step, we present visual examples in \figureautorefname{ \ref{fig:visualization}}.
The second-column visual examples show that our baseline obviously address the issue of spatial inconsistency. However, examples in column (b) appear severe unclear boundaries between different facial components in the first three green boxes, and confusing contours of multi-faces in the last three green boxes. This is due to the fact that the baseline lacks a reasoning ability on global dependencies. 
The first three-row examples in column (c) show complete structure of individual component and almost clear boundaries between facial parts, illustrating the long-range inference ability of our proposed DDGCN. Nonetheless, the last three-row examples in the column (c) still exist different face instances, as the proposed DDGCN has a limited capability of localizing objects of similar contours. 
Compared to examples in columns (b)-(c), columns (d)-(e) present clear boundaries between different facial components in both single-face and multi-face scenes, due to the feature enhancement by semantic edges. 
Looking at the areas within green rectangles in columns (d)-(e), CSR can recover error pixels, preventing noisy labels in training datasets from degrading model generalization.

\section{Conclusion}

In this paper, we present DML-CSR, a decoupled multi-task learning method with cyclical self-regulation for face parsing. Comprehensive experiments on Helen, CelebAMask-HQ, and LaPa verify the effectiveness of the proposed method. The results show that DML-CSR significantly outperforms other methods on all datasets. Training details will be released to encourage further research towards face parsing.

\noindent\textbf{Limitations.} Our method achieves impressive results in face parsing. However, there is a slight performance degradation in low-resolution faces. This is because that we train our model on the high-resolution face dataset. Even so, we believe DML-CSR is a valuable method for training a reliable face parsing model on a large-scale dataset.

\noindent\textbf{Societal Impact.} We develop a general model for face parsing in this paper, and the proposed model is not used for a specific application. Therefore, this work does not directly involve societal issues.

\noindent\textbf{Acknowledgements.} Stefanos Zafeiriou acknowledges support from the EPSRC Fellowship DEFORM (EP/S010203/1), FACER2VM (EP/N007743/1) and a Google Faculty Fellowship. 
Ying Li acknowledges support from the National Natural Science Foundation of China (61871460).

{\small
\bibliographystyle{ieee_fullname}
\bibliography{egbib}
}

\end{document}